\RequirePackage{silence}
\RequirePackage{fix-cm}
\documentclass[10pt,twocolumn]{ICCAS}
\usepackage{graphics} % for pdf, bitmapped graphics files
\usepackage{epsfig} % for postscript graphics files
\usepackage{mathptmx} % assumes new font selection scheme installed
\usepackage{times} % assumes new font selection scheme installed
\usepackage{amsmath} % assumes amsmath package installed
\usepackage{amssymb}  % assumes amsmath package installed
\usepackage{kotex}
\usepackage{tabularx}
\usepackage{booktabs}
\usepackage{graphicx}
\usepackage{algorithm}
\usepackage{algpseudocode}
\usepackage{subcaption}
\usepackage[font=footnotesize,labelfont=bf]{caption}
\usepackage{placeins}
\usepackage{afterpage}
\usepackage{multirow}
\usepackage{comment}
\usepackage{cite}
\usepackage{url}
\usepackage{hyperref}
\usepackage{xcolor}
\usepackage{array}
\usepackage{makecell}
\usepackage{pifont}
\usepackage{listings}
\makeatletter
\@ifundefined{c@paragraph}{\newcounter{paragraph}[subsubsection]}{}
\@ifundefined{theparagraph}{}{}
\makeatother

\begin{document}

\title{Spatial and Semantic Reasoning for LLM-Driven Robot Navigation via MCP}

\author{Jungsoo Lee$^{1,\dagger}$, Jaegyun Park$^{1,\dagger}$ and Wansoo Kim$^{2,*}$}

\affils{ ${}^{1}$Department of Robotics, Hanyang University, \\
Seoul, 04763, Korea (\{lpigeon, chant29\}@hanyang.ac.kr) \\
${}^{2}$Department of Robotics, Hanyang University ERICA, \\
Ansan, 15588, Korea (wansookim@hanyang.ac.kr) \\
{\small${}^{\dagger}$ These authors contributed equally to this work.}
{\small${}^{*}$ Corresponding author}
}

%\thanks{ \noindent
%   This paper is supported by my funding agencies.
%  }

\abstract{
   Large language models (LLMs) are increasingly used as natural-language interfaces for robotic systems, yet their integration with Robot Operating System (ROS)-based navigation remains limited by two gaps. First, navigation data such as occupancy grids are represented as raw geometric messages that are difficult for LLMs to use directly as spatial or semantic context. Second, adding LLM-driven capabilities often requires custom wrappers or robot-specific interfaces, limiting reuse across systems. To address these challenges, we propose a non-invasive framework that connects LLM reasoning with ROS-based navigation through a navigation-oriented representation layer, exposed through the Model Context Protocol (MCP) as standardized, reusable tools so that any MCP-compatible LLM can access them without robot-specific wrappers. The visual map modules transform occupancy grids into metric, pose-aware images for goal reasoning, while the semantic annotation modules record waypoint-level observations with robot poses. We evaluate the framework on three tasks: autonomous mapping, spatial reasoning-based navigation, and semantic reasoning-based navigation. The results show that the evaluated LLM backends use these representations to achieve over 97\% map coverage and select spatial or semantic navigation targets from natural-language instructions in a simulated indoor environment. This demonstrates representation-mediated LLM navigation without modifying the existing ROS navigation stack.
}


\maketitle

%-----------------------------------------------------------------------

\section{Introduction}
Humans can flexibly coordinate complex tasks through natural language and modify actions in real time. If robotic systems could be endowed with such capabilities, even non-expert users would be able to intuitively control robots in dynamic environments. Recently, large language models (LLMs), including their multimodal variants, have emerged as powerful high-level controllers and conversational interfaces for realizing this vision~\cite{zhang2023large, wang2025large}.

Despite these advances, a gap persists between LLMs and physical robotic systems, especially for autonomous navigation in the Robot Operating System (ROS) ecosystem. First, ROS navigation map data, such as occupancy grids, are typically stored as arrays of per-cell numerical occupancy values, making them difficult for LLMs to interpret directly as spatial or semantic context without an appropriate representation~\cite{xie2024empowering}. Second, adding new LLM-driven robot capabilities often requires custom wrappers or robot-specific interfaces, limiting extensibility across systems.

Existing LLM--robot interfaces make robot functions accessible to language models, but they do not necessarily define navigation-specific representations for map, pose, and perception data. Conversely, many LLM-based navigation approaches rely on pre-built or separately constructed spatial and semantic representations~\cite{lmnav,vlmaps,l3mvn}, which differs from directly augmenting an existing ROS navigation stack with external representation tools.

These limitations highlight the need to bridge raw ROS navigation data and LLM-based goal reasoning without redesigning the underlying robot software.

To address these challenges, this paper proposes a non-invasive framework that connects LLM reasoning with ROS-based navigation through a navigation-oriented representation layer. Rather than treating the LLM as a command translator or simply exposing raw ROS messages, the framework converts map, pose, and perception data into spatial and semantic contexts that the LLM can use for goal reasoning. Through this representation layer, the LLM infers exploration and navigation goals from visual map structure and waypoint-level semantic observations while the underlying robot software stack remains unchanged.

\begin{enumerate}
\setlength{\itemsep}{0pt}
\setlength{\parsep}{0pt}
\item \textbf{Navigation-oriented visual map representation:} Rather than embedding map reasoning in a dedicated navigation pipeline, we design a visual map representation that converts occupancy-grid messages, originally stored as numerical cell values, into metric, pose-aware map images and exposes them as standardized external tools. This provides the LLM with spatial cues for reasoning about environmental layout, unexplored regions, and candidate navigation goals on top of an unmodified ROS stack.
\item \textbf{Waypoint-centric semantic grounding:} We define a semantic data specification and annotation process that associates natural-language descriptions of observed landmarks, geometric features, and places with the robot pose at each exploration waypoint, enabling object- or place-related user instructions to approximate, waypoint-level navigation targets rather than exact object coordinates.
\item \textbf{Non-invasive integration with ROS navigation:} We implement the representation layer as standardized external tools connected to existing ROS SLAM and navigation stacks, demonstrating autonomous mapping and reasoning-based navigation across two LLM backends without modifying the underlying robot-side ROS SLAM or navigation stack.
\end{enumerate}

\section{RELATED WORK}

LLM--ROS integration frameworks have improved robot accessibility by exposing ROS functions to language models. ROSA~\cite{rosa2024} wraps ROS topics and services as LangChain tools, ROS-LLM~\cite{mower2024rosllm} generates behavior trees and state machines from ROS actions and services, RoboNeuron~\cite{roboneuron2024} translates ROS~2 message structures into MCP tools, and ROS-MCP Server~\cite{rosmcp} exposes ROS topics, services, actions, and parameters through MCP. These systems reduce integration effort and enable LLMs to inspect or command robot systems through reusable interfaces. However, they mainly address access to robot functions, rather than how raw navigation data should be represented as spatial and semantic context for LLM-based goal reasoning.

LLM- and VLM-based navigation studies have addressed spatial and semantic grounding from a complementary direction. Prior methods use pre-built topological graphs~\cite{lmnav}, semantic/frontier-based target-navigation representations such as L3MVN and VLFM~\cite{l3mvn,vlfm}, occupancy-grid structural extraction~\cite{espada2025leveraging}, text-based semantic maps~\cite{xie2025intelligent}, or semantic-spatial memory retrieval~\cite{meta_memory}. While effective for language-guided navigation and search, these representations are often coupled with method-specific navigation components, such as topological routing, frontier selection, or memory retrieval, rather than exposed as reusable tools for an existing ROS navigation stack. 

In contrast, our framework focuses on the representation gap between generic ROS access and LLM-based navigation reasoning, rather than on a task-specific navigation pipeline. It keeps the existing SLAM and navigation stack unchanged, while exposing visual map representations and waypoint-centric semantic records as MCP-compatible external tools. Table~\ref{table_comparison} summarizes this distinction across protocol standardization, ROS compatibility, navigation-oriented representation, and robot-side stack modification.

% In contrast, our framework focuses on the representation gap between generic ROS access and LLM-based navigation reasoning. It keeps the existing SLAM and navigation stack unchanged while providing real-time visual map representations and waypoint-centric semantic records as standardized external tools. 

\begin{figure*}[!t]
\centering
\includegraphics[width=1.0\linewidth]{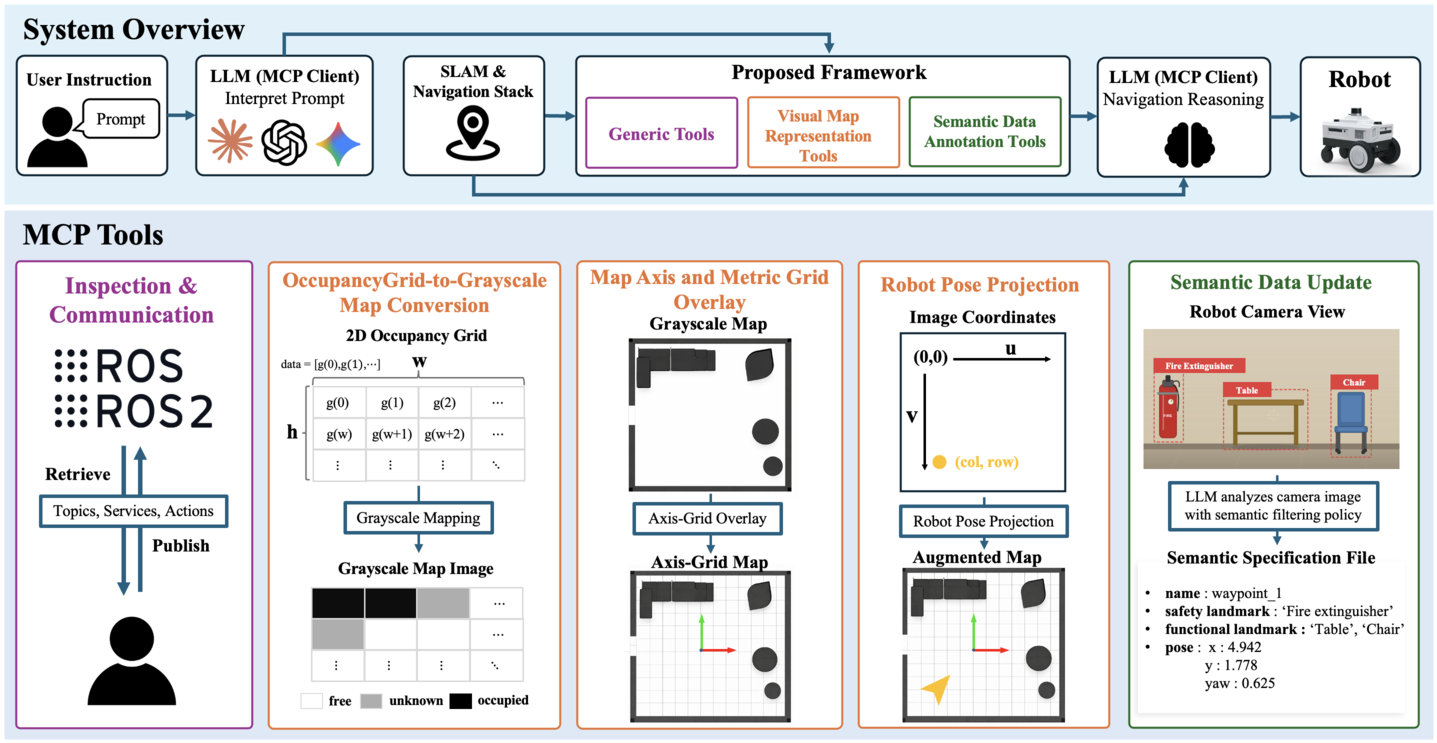}
\caption{System overview and MCP tool composition of the proposed framework. (Top) A user's natural-language command is interpreted by an LLM, grounded through the proposed representation tools, and executed by the existing ROS SLAM and navigation stack. (Bottom) The MCP tools provide generic ROS inspection and communication, visual map representation from occupancy grids, and semantic data annotation from camera observations with robot poses.}
\label{fig:framework}
\end{figure*}
\section{METHODOLOGY}
\label{sec:methodology}

\subsection{System Overview}

The Model Context Protocol~(MCP) standardizes how LLM applications access external tools and data sources through client--server interfaces~\cite{mcp_hou2025}. The proposed framework uses MCP as the common tool interface between an LLM client and an unmodified ROS-based SLAM and navigation stack. ROS-MCP Server~\cite{rosmcp} is used as the generic communication substrate for accessing ROS topics, services, actions, and parameters. The proposed framework extends this substrate with representation modules that convert navigation-relevant ROS data, including occupancy-grid map data, robot poses, and camera observations, into LLM-interpretable spatial and semantic contexts. Each representation module is exposed as an MCP tool by decorating a Python function with \texttt{@mcp.tool()} and declaring a JSON schema, after which it can be invoked by any MCP-compatible client. The overall system architecture and the proposed modules are illustrated in Fig.~\ref{fig:framework}.

\newcommand{\cmarkk}{\large$\checkmark$}
\newcommand{\xmarkk}{\large$\times$}
\newcommand{\tmarkk}{\large$\triangle$}

\begin{table}[t]
\caption{Comparison of LLM-ROS Integration Frameworks.}
\label{table_comparison}
\centering
\scriptsize
\renewcommand{\arraystretch}{1.5}
\setlength{\tabcolsep}{0.6pt}
\noindent\resizebox{0.98\columnwidth}{!}{%
\begin{tabular}{>{\raggedright\arraybackslash}m{2.15cm}cccccc}
\hline
\textbf{Feature} & \textbf{\,ROSA\,} & \textbf{\,ROS-LLM\,} & \textbf{\,RoboNeuron\,} & \textbf{\,Espada \textit{et al.}\,} & \makecell{\textbf{ROS-MCP}\\\textbf{Server}} & \textbf{\,Ours\,} \\
\hline\hline
\makecell[l]{Standardized\\protocol (MCP)} & \xmarkk & \xmarkk & \cmarkk & \xmarkk & \cmarkk & \cmarkk \\
\hline
\makecell[l]{ROS version-\\agnostic} & \cmarkk & \xmarkk & \xmarkk & \xmarkk & \cmarkk & \cmarkk \\
\hline
\makecell[l]{Navigation-\\oriented\\representation} & \xmarkk & \xmarkk & \xmarkk & \tmarkk$^{a}$ & \xmarkk & \cmarkk \\
\hline
\makecell[l]{No robot-side\\stack modification} & \tmarkk$^{b}$ & \tmarkk$^{b}$ & \cmarkk & \tmarkk$^{b}$ & \cmarkk & \cmarkk \\
\hline
\end{tabular}
}

\begin{minipage}{0.98\columnwidth}
\raggedright\scriptsize
$^{a}$Task-specific navigation representation rather than a reusable external representation layer.\\
$^{b}$No direct ROS-stack patching, but robot- or task-specific external modules may be required.
\end{minipage}
\end{table}

\subsection{Visual Map Representation Tools}

ROS navigation maps are commonly provided as \texttt{nav\_msgs/OccupancyGrid} messages, whose raw numerical arrays are inefficient for LLM-based reasoning when directly serialized as text. Because the size of an occupancy-grid array increases with map resolution and spatial extent, directly passing grid values to an LLM can introduce unnecessary input complexity and latency while still providing limited geometric structure for spatial interpretation. To address this issue, the proposed visual map representation module converts the occupancy grid into a coordinate-aware image that preserves the geometric structure of the environment while providing metric references for goal reasoning. The \texttt{subscribe\_map\_as\_image} tool reshapes occupancy values into a 2D grid and maps each cell value $g\in\{-1,0,1,\dots,100\}$ to grayscale intensity $I(g)$ as
\begin{equation}
I(g)=
\begin{cases}
127, & g=-1, \\
255, & g=0, \\
255\left(1-\frac{g}{100}\right), & 1 \le g \le 100.
\end{cases}
\label{eq:occupancy_conversion}
\end{equation}
Here, unknown cells are gray, free cells are white, and occupied cells become darker as occupancy probability increases. The generated image is vertically flipped to align the ROS map coordinate convention with the image coordinate frame.

The \texttt{draw\_map\_axis} tool overlays coordinate axes and an adaptive metric grid using the map width $W$, height $H$, resolution $r$, and origin $(x_o,y_o)$. The physical map size is
\begin{equation}
L_x = Wr, \qquad L_y = Hr.
\label{eq:map_size}
\end{equation}
A world coordinate $(x,y)$ is projected to image coordinates $(u,v)$ by
\begin{equation}
u = \frac{x-x_o}{r}, \qquad
v = H - \frac{y-y_o}{r},
\label{eq:transform_to_imageframe}
\end{equation}
and the world origin is located at
\begin{equation}
u_0 = \frac{-x_o}{r}, \qquad
v_0 = H - \frac{-y_o}{r}.
\label{eq:origin_in_imageframe}
\end{equation}
To avoid visual clutter, the grid spacing is selected from intuitive metric values after computing
\begin{equation}
s_{\mathrm{base}} = \max\left(\frac{L_s}{N},r\right), \qquad L_s=\min(L_x,L_y), \quad N \approx 15.
\label{eq:base_spacing}
\end{equation}

The \texttt{draw\_robot\_pose} tool projects the current robot pose $(x,y,\theta)$ onto the map as an arrow marker. For each local arrow point $(l_x,l_y)$ in the robot frame, the image-frame displacement is
\begin{equation}
\begin{aligned}
\Delta col &= l_x \cos\theta - l_y \sin\theta, \\
\Delta row &= -(l_x \sin\theta + l_y \cos\theta).
\end{aligned}
\label{eq:arrow_geometry}
\end{equation}
Together, the map image integrates environment structure, metric scale, and robot state, allowing the LLM to infer unexplored regions, relative room sizes, and feasible navigation goals.

\subsection{Semantic Data Annotation Tools}

The visual map representation provides geometric context, while semantic reasoning-based navigation requires additional semantic context that complements geometry by associating natural-language referents with observed objects, places, and functional cues in the environment. The proposed semantic data annotation module stores waypoint-level observations in a YAML-based specification file together with the robot pose at which they were observed.

To reduce unstable or ambiguous semantic entries, the LLM follows a prompt-defined semantic filtering policy. This policy is not treated as confidence calibration or quantitative reliability estimation; instead, it serves as a qualitative, instruction-level guideline for retaining observations that are likely to be stable, recognizable, and useful for navigation.

The semantic filtering policy organizes retained observations into safety landmarks, functional landmarks, and geometric features. These categories cover fixed safety-related equipment, fixed task-relevant equipment or objects, and persistent geometric layout cues such as corridors, walls, openings, and obstacle arrangements.

Each semantic waypoint is represented as
\begin{equation}
s_i =
\{n_i, d_i, p_i,
L_i^{safe}, L_i^{func}, F_i^{geom}\},
\label{eq:semantic_location}
\end{equation}
where $n_i$ is the waypoint name, $d_i$ is the natural-language place description, and $p_i=(x_i,y_i,\theta_i)$ is the robot observation pose in the map frame. The sets $L_i^{safe}$, $L_i^{func}$, $F_i^{geom}$ denote safety landmarks, functional landmarks, geometric features, respectively.

The complete semantic map is represented as
\begin{equation}
S = \{s_1, s_2, \dots, s_N\},
\label{eq:semantic_map}
\end{equation}
where $N$ is the number of stored semantic locations.

The \texttt{write\_semantic\_data} tool creates or updates this specification at each exploration waypoint after the LLM analyzes the robot camera observation together with the annotated map image and applies the semantic filtering policy. During navigation, the \texttt{get\_semantic\_data} tool retrieves the stored records, and the LLM identifies the semantic waypoint most relevant to the requested object, function, or place. The associated pose $p_i$ is then used as an approximate navigation goal, grounding instructions to waypoint-level semantic anchors rather than exact object coordinates.

\subsection{Execution Flow}
The user provides a natural-language instruction, which is processed by the LLM through an MCP client. Generic ROS interface tools are used to access the SLAM and navigation stack, while the proposed representation modules convert raw occupancy grids into annotated map images and maintain waypoint-level semantic records from camera observations. The LLM synthesizes these spatial and semantic contexts, determines an appropriate goal position, and issues a navigation goal through the existing navigation stack for path planning, obstacle avoidance, and trajectory tracking.

\section{EXPERIMENTS}
\label{sec:experiments}

\subsection{Experimental Setup}

\begin{figure}[t]
    \centering
    \includegraphics[width=1.0\linewidth]{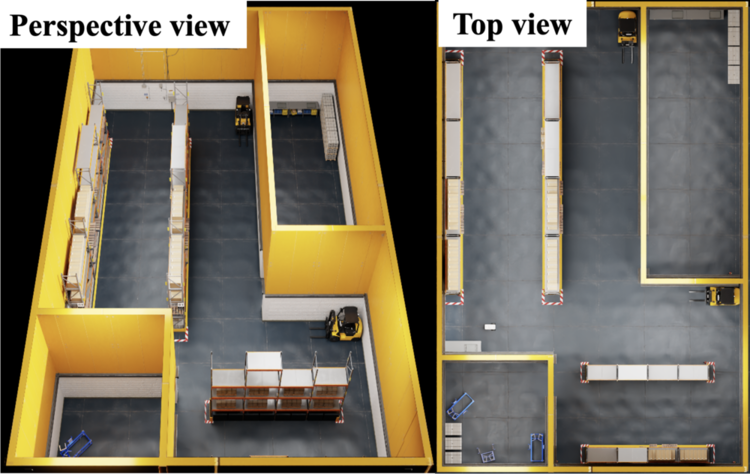}
    \caption{Experimental setup in NVIDIA Isaac Sim. The indoor warehouse environment is shown in perspective view (left) and top view (right), and includes two rooms of different sizes, four shelves loaded with goods, two forklifts, and safety equipment such as fire extinguishers.}
    \label{fig:env}
\end{figure}

The experiment is conducted in the NVIDIA Isaac Sim simulator using a Nova Carter robot. The robot is equipped with a LiDAR sensor and operates on a ROS\,2-based software stack. Cartographer is used for SLAM and Nav2 for navigation. The experimental environment consists of an indoor warehouse space with two rooms and shelves, where various objects are placed throughout. The layout of the experimental setup is shown in Fig.~\ref{fig:env}. The LLM backends used in this study are Claude Sonnet 4.6 and GPT-5.5, both accessed through the MCP interface. To guide efficient exploration, a predefined prompt is provided to the LLM at the start of the exploration task; the full prompt is given in the Appendix.

\subsection{Task Design}

To validate the proposed framework, we design an experimental scenario consisting of three tasks. Task~1 is first performed to generate an annotated map and semantic data, which are then used as input to evaluate Task~2 and Task~3 independently. The prompts used for each task are presented in Table~\ref{tab:scenarios}.

\subsubsection{Task 1 -- Autonomous Mapping and Semantic Data Collection}

The user instructs the robot, through natural language, to explore the environment and record semantic observations. The LLM uses the visual map representation tools to convert the occupancy grid into an annotated map image and identify unexplored regions. It then performs spatial reasoning based on the information accumulated from previous waypoints to select an unexplored region suitable for the next stage of exploration as the next waypoint. The robot navigates to the selected location and captures a camera image. The LLM then uses the semantic data annotation tools to record selected semantic observations, including visual landmarks from the camera observation and geometric layout cues inferred from the annotated map image, together with the current waypoint pose in the semantic data specification file. This process is repeated until the LLM determines, from the visual map representation, that the exploration coverage has reached 90\%.

\subsubsection{Task~2 -- Spatial Reasoning-Based Navigation}

The user issues a navigation goal requiring spatial reasoning. The LLM uses map-derived spatial properties from the annotated map image, while referring to semantic data when needed, to determine a goal position satisfying the given condition. The robot then navigates to the determined position through the navigation stack.

\subsubsection{Task~3 -- Semantic Reasoning-Based Navigation}

The user instructs the robot to navigate to a specific object using an abstract expression. The LLM infers the object intended by the user from semantic entries in the semantic data specification file and uses the associated waypoint pose as an approximate navigation target. The robot then navigates to the determined position through the navigation stack.

\begin{figure*}[!t]
\centering
\includegraphics[width=0.9\linewidth]{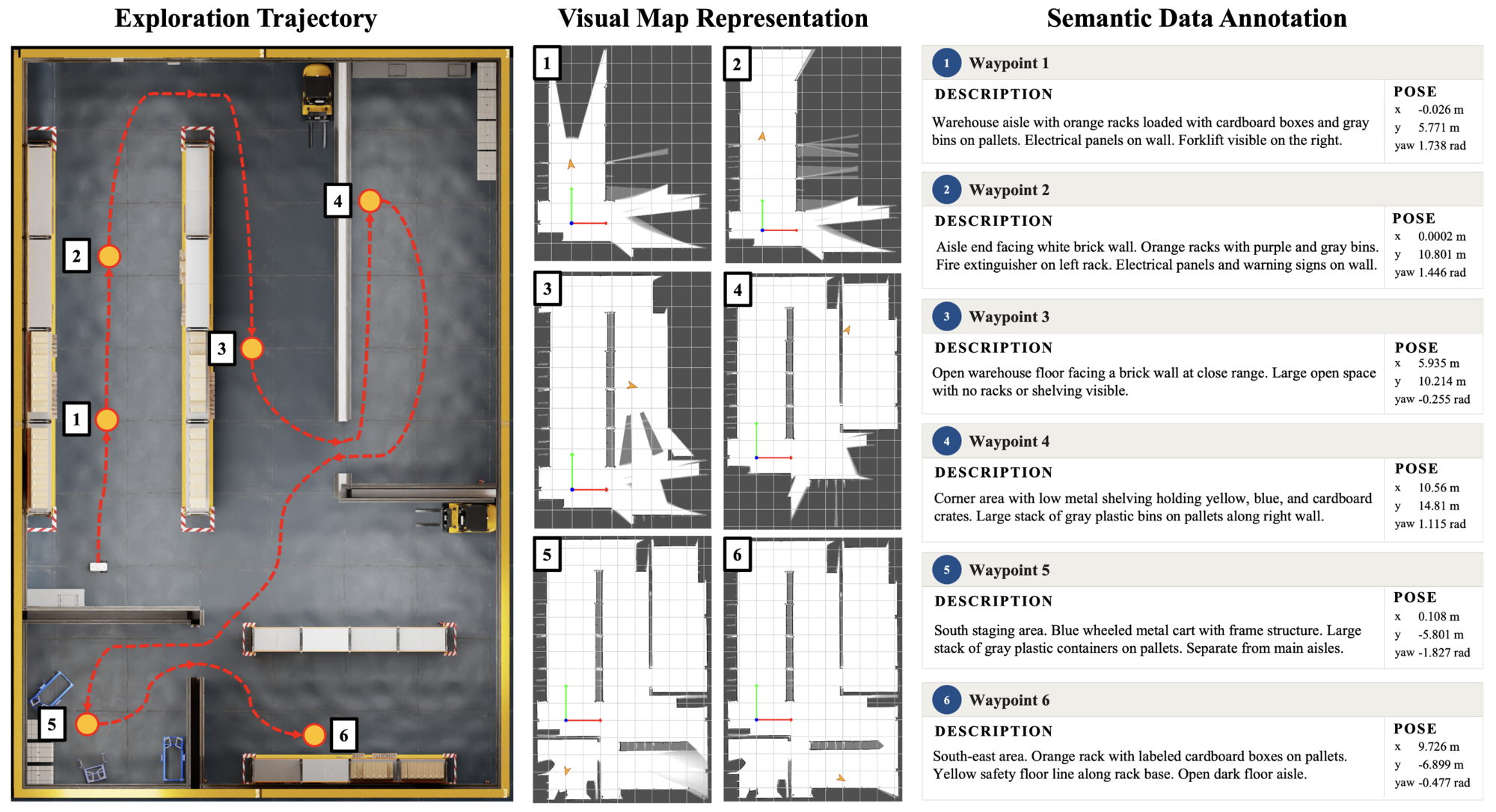}
\caption{Representative Claude Sonnet 4.6 execution results of Task~1. (Left) Exploration trajectory over six LLM-generated waypoints executed by Nav2. (Center) Annotated map images generated by the visual map representation tools at each waypoint, showing progressive map coverage during exploration. (Right) Waypoint-level semantic observations recorded by the semantic data annotation tools with pose information.}
\label{fig:results_1}
\end{figure*}

\begin{table}[t]
\caption{User prompts for each evaluation task.}
\label{tab:scenarios}
\centering
\footnotesize
\renewcommand{\arraystretch}{1.25}
\begin{tabular}{cp{6.5cm}}
\hline
\textbf{Task} & \textbf{Prompt} \\
\hline\hline
1 & ``Update the map while moving.'' \\
\hline
2 & ``Go to the smallest room.'' \\
\hline
3 & ``Fire detected nearby. Navigate to the tool used for it.'' \\
\hline
\end{tabular}
\end{table}

\subsection{Evaluation}

For each LLM backend, Task~1 was executed independently to generate a backend-specific annotated map and semantic data, which were then used for the corresponding Task~2 and Task~3 evaluations.

%Task~1 is evaluated based on whether the environment is mapped with a coverage rate of at least 90\%, defined as the ratio of free-space pixels in the generated map to those in the ground truth map constructed via manual teleoperation, and semantic observations are recorded in the semantic data specification file. 
Task~1 is evaluated based on two criteria: first, whether the environment is mapped with a coverage rate of at least 90\% (defined as the ratio of free-space pixels in the generated map to those in a ground-truth map constructed via manual teleoperation); and second, whether semantic observations are correctly recorded in the semantic data specification file.
Task~2 and Task~3 are each repeated 30 times. Task~2 measures the number of successful trials in which the robot navigates to a goal satisfying the spatial condition (the smallest room), and Task~3 measures the number of successful trials in which the robot navigates to the waypoint associated with the target object (the fire extinguisher).

\begin{figure}[t]
    \centering
    \includegraphics[width=1.0\linewidth]{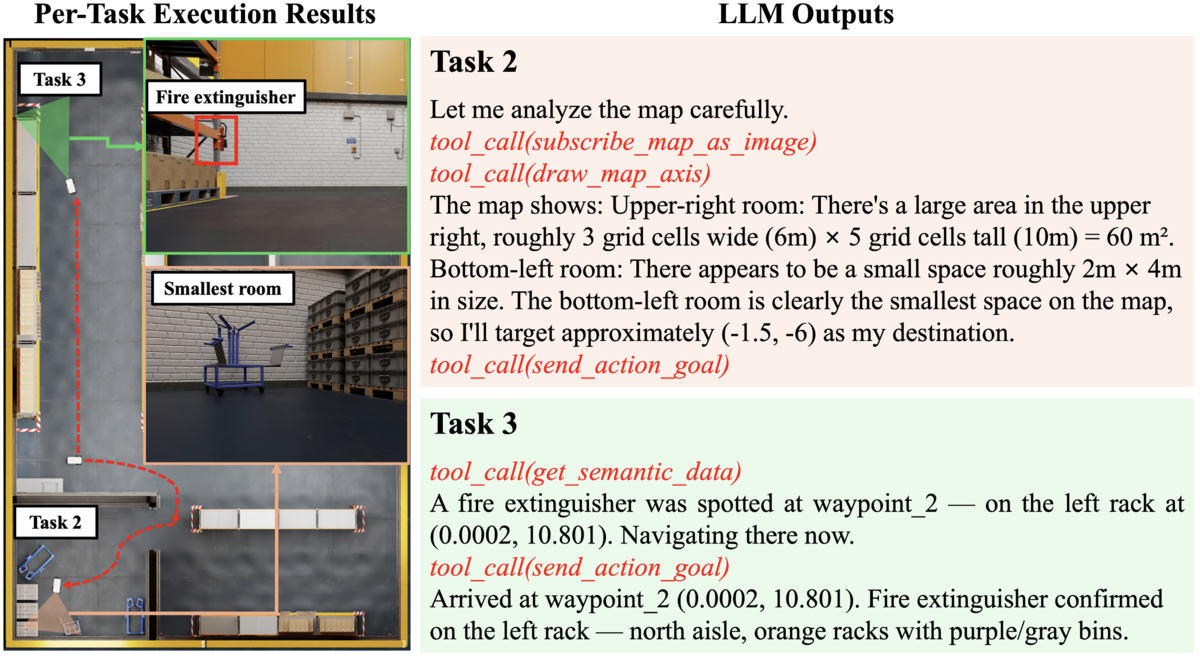}
    \caption{Representative Claude Sonnet 4.6 results for Task~2 and Task~3. (Left) Robot trajectories and destination camera images on the top-view environment: Task~2 reaches the smallest room via spatial reasoning, and Task~3 selects the fire-extinguisher waypoint from the semantic data. (Right) LLM tool-call sequences and reasoning outputs.}
    \label{fig:results_2}
\end{figure}

\subsection{Results}

Fig.~\ref{fig:results_1} illustrates Task~1, and Fig.~\ref{fig:results_2} illustrates Task~2 and Task~3, while Table~\ref{tab:llm_results} summarizes the task results for both LLM backends.

Task~1 demonstrates that the LLM autonomously explored the environment using the proposed visual map representation tools, iteratively identifying unexplored regions and expanding map coverage. Claude Sonnet 4.6 and GPT-5.5 achieved coverage rates of 97.83\% and 99.78\%, respectively. Additionally, during the exploration process, selected semantic observations recognized from camera images and the corresponding waypoint locations were recorded together in the semantic data specification file.

Task~2 demonstrates that, given a user prompt, the LLM interpreted the annotated map image, estimated relative room sizes based on grid-cell counts, and determined the goal position satisfying the given condition. Claude Sonnet 4.6 succeeded in 26 out of 30 trials (86.67\%), while GPT-5.5 succeeded in 28 out of 30 trials (93.33\%). Failures occurred when the LLM misidentified a corridor as a room or incorrectly estimated room sizes, leading to erroneous spatial reasoning and incorrect goal selection.

Task~3 demonstrates that the LLM interpreted the user's abstract prompt, retrieved semantic entries associated with the target object from the semantic data specification file, and determined the corresponding waypoint pose as the navigation goal. When multiple entries containing the requested object were present, it additionally retrieved the robot's current pose and selected the nearest associated waypoint as the navigation target. Both LLM backends succeeded in 30 out of 30 trials (100\%), as the presence of target-object entries in the semantic data specification file enabled consistent goal selection in the evaluated environment.

\begin{table}[t]
\caption{Task results by LLM backend.}
\label{tab:llm_results}
\centering
\footnotesize
\renewcommand{\arraystretch}{1.2}
\resizebox{\columnwidth}{!}{%
\begin{tabular}{lccc}
\hline
\textbf{LLM Backend} & \textbf{Task~1} & \textbf{Task~2} & \textbf{Task~3} \\
\hline\hline
Claude Sonnet 4.6 & 97.83\% coverage & 26/30 (86.67\%) & 30/30 (100\%) \\
GPT-5.5 & 99.78\% coverage & 28/30 (93.33\%) & 30/30 (100\%) \\
\hline
\end{tabular}
}
\end{table}

\vspace{-1.2em}

\section{DISCUSSION}
\label{sec:discussion}

The proposed framework shows that a multimodal LLM can perform mapping and navigation from spatial and semantic representations without dedicated reasoning modules, but its performance depends on both representation quality and backend capability. Task~2 failures indicate sensitivity to ambiguous spatial structures and backend-dependent spatial interpretation, while Task~3 should be interpreted as navigation to waypoint-level semantic anchors under complete annotations rather than exact object-coordinate navigation. Future work will improve map representation and semantic annotation, and investigate a robust grounding layer that combines semantic retrieval with spatial consistency checking under incomplete or noisy observations.

\section{CONCLUSION}
\label{sec:conclusion}

This study proposes a non-invasive LLM-based navigation framework that connects ROS, LLM reasoning, and navigation through visual map representation and semantic data annotation modules. By enabling the LLM to interpret spatial structures and ground natural-language referents to map locations, the framework supports autonomous mapping, spatial reasoning-based navigation, and semantic reasoning-based navigation without modifying the existing ROS infrastructure. These results indicate the feasibility of representation-mediated LLM navigation beyond simple command generation.

\vspace{-1.2em}

\section*{APPENDIX : EXPLORATION PROMPT}
\label{app:prompt}
The predefined prompt provided to the LLM at the start of the exploration task is shown below.

\begin{lstlisting}[basicstyle=\ttfamily\scriptsize, breaklines=true, breakatwhitespace=true, columns=fullflexible]
Repeat at every waypoint:
Step 1 - Map Analysis & Waypoint Selection
- Use visual observations from the previous waypoint to determine the next exploration target.
- Select ONE waypoint >= 5 m away toward the largest unexplored area. No backtracking within 3 m of visited locations.
- Set goal yaw facing the unexplored or most informative direction at that location.
Step 2 - Navigation
- Send navigation goal and monitor until success.
Step 3 - Semantic Recording
- Capture front camera image and analyze visible landmarks, geometric features, and task-relevant objects.
- Save semantic data according to the semantic map policy via write_semantic_data at the current (x, y, yaw).
Step 4 - Repeat until at least 90% coverage.
\end{lstlisting}

\vspace{-1.2em}

\section*{ACKNOWLEDGEMENT}
This work was supported by the National Research Foundation of Korea (NRF) grant funded by the Korean government (Ministry of Science and ICT) (No. 2022R1C1C1008306) and by the Alchemist Project (RS-2024-00432143) funded by the Ministry of Trade, Industry and Energy (MOTIE), Korea.

% This work was supported by the National Research Foundation of Korea (NRF) grant funded by the Korea government (Ministry of Science and ICT) (No. 2022R1C1C1008306).
% This work was supported by the Alchemist Project (RS-2024-00432143) funded by the Ministry of Trade, Industry & Energy (MOTIE, Korea).
%This paper has been supported by NRF of Korea in 2021.

%%%%%%%%%%%%%%%%% BIBLIOGRAPHY IN THE LaTeX file !!!!! %%%%%%%%%%%%%%%%%%%%%%
%%---------------------------------------------------------------------------%%
%
\bibliographystyle{IEEEtran}
\bibliography{references}
%
%%--------------------------------------------------------------------%%

\end{document}